\documentclass[11pt,a4paper]{article}

\usepackage[utf8]{inputenc}
\usepackage[T1]{fontenc}
\usepackage[a4paper,margin=1in]{geometry}

\usepackage{times}            % Standard in many ML venues
\usepackage{microtype}        % Better kerning and justification

\usepackage{amsmath,amssymb,amsfonts}

\usepackage{graphicx}
\usepackage{booktabs}
\usepackage{caption}
\usepackage{authblk}

\usepackage{hyperref}
\usepackage[nameinlink,capitalise,noabbrev]{cleveref}

\hypersetup{
    colorlinks=true,
    linkcolor=blue,
    citecolor=blue,
    urlcolor=magenta,
    pdftitle={Pretraining Transformer-Based Models on Diffusion-Generated Synthetic Graphs for Alzheimer's Disease Prediction},
    pdfauthor={Abolfazl Moslemi, Hossein Peyvandi},
}

\title{\textbf{Pretraining Transformer-Based Models on \\ Diffusion-Generated Synthetic Graphs for \\ Alzheimer's Disease Prediction}}

\author[1]{\textbf{Abolfazl Moslemi}}
\author[2]{\textbf{Hossein Peyvandi}}
\affil[1]{Sharif University of Technology, Tehran, Iran\\
\texttt{moslemiabolfazl14@gmail.com}}
\affil[2]{Pasargad Institute for Advanced Innovative Solutions\\
\texttt{h.peyvandi@sharif.ir}}

\date{22 Nov 2025} % No date is displayed if you use \date{} instead

\begin{document}

\maketitle

\begin{abstract}
\noindent Early and accurate detection of Alzheimer's disease (AD) is crucial for enabling timely intervention and improving patient outcomes. However, developing reliable machine learning (ML) models for AD diagnosis remains challenging due to limited labeled biomedical data, multi-site heterogeneity, and class imbalance. To address these constraints, we propose a Transformer-based diagnostic framework that integrates diffusion-based synthetic data generation with graph representation learning and transfer learning. Specifically, we employ a class-conditional denoising diffusion probabilistic model (DDPM) trained on the real-world NACC dataset to generate a large-scale synthetic cohort that mirrors the multimodal clinical and neuroimaging feature distributions of AD patients while balancing diagnostic classes. Modality-specific Graph Transformer encoders are first pretrained independently on this synthetic data to learn robust, class-discriminative representations. In the downstream stage, we freeze these pretrained encoders and train a neural classifier on top of the resulting multimodal embeddings using the original NACC data.

We further quantify the distributional alignment between real and diffusion-generated cohorts using metrics such as Maximum Mean Discrepancy (MMD), Fr\'echet distance, and energy distance, and complement standard discrimination metrics with calibration analysis, sensitivity at fixed specificity, decision curve analysis, and subgroup evaluations (e.g., stratified by age, sex, and APOE4 status) to probe clinical utility. Empirically, our framework outperforms standard baselines, including early and late fusion deep neural networks and the multimodal graph-based model MaGNet, achieving higher AUC, accuracy, sensitivity, and specificity under subject-wise cross-validation on NACC. By leveraging generative modeling for domain-aligned pretraining and using Graph Transformers as modality-specific feature extractors, our method improves generalization in low-sample, imbalanced settings and highlights the potential of diffusion-based synthetic pretraining for data-efficient clinical prediction models in neurodegenerative disease diagnosis.
\end{abstract}

\section{Introduction}
Alzheimer's disease (AD) is a progressive neurodegenerative disorder and the most common cause of dementia in older adults, affecting millions globally. Despite advances in understanding its pathology and the recent development of disease-modifying therapies \cite{vandyck2023}, early and accurate diagnosis remains a significant clinical challenge. Therapeutic interventions are most effective in the early stages of AD, yet current diagnostic tools are often invasive (e.g., lumbar punctures), costly (e.g., PET scans), or lack sufficient sensitivity and specificity \cite{knopman2021}. This has motivated the growing use of machine learning (ML) and artificial intelligence (AI) to develop non-invasive, multimodal predictive models for early AD detection.

However, a major obstacle in developing robust ML models in the biomedical domain is the scarcity of labeled data. This issue is particularly pronounced in neurodegenerative disease cohorts, where data collection is expensive, multi-site harmonization is challenging, and class imbalance is common. Moreover, real-world datasets such as the National Alzheimer's Coordinating Center (NACC) \cite{beekly2004, weintraub2009} aggregate measurements across multiple clinical sites, introducing domain shifts that complicate generalization. To address these issues, recent advances in generative modeling have opened new avenues. In particular, denoising diffusion probabilistic models (DDPMs) \cite{ho2020, karras2022} have emerged as powerful tools for generating high-fidelity, structured synthetic data that can augment limited datasets and support pretraining in data-scarce regimes.

In this work, we propose a framework for AD diagnosis that integrates synthetic data generation with graph-based representation learning and transfer learning. We first train a class-conditional DDPM on real data from NACC to generate a large, balanced synthetic dataset that mimics the joint distribution of multimodal features---specifically, clinical data from the Uniform Data Set (UDS) and graph-structured MRI-derived measures. These synthetic samples are then used to pretrain modality-specific Graph Transformer encoders that serve as feature extractors. After pretraining, we construct the full architecture by freezing the modality-specific encoders and training a downstream classifier on top of the resulting multimodal embeddings using the real Alzheimer dataset in a second stage.

From a geometric perspective, the DDPM learns score-based flows that approximate class-conditional data manifolds, while the Graph Transformers learn functions on graphs constructed from the same feature spaces. This suggests a natural synergy: synthetic samples generated along the learned manifolds provide diverse yet domain-aligned inputs that allow Graph Transformers to explore a broader range of neighborhood configurations before being adapted to the limited real cohort. We make this connection more explicit by introducing a theoretical motivation for diffusion-based synthetic pretraining and by quantifying synthetic--real distribution shift using metrics such as Maximum Mean Discrepancy (MMD), Fr\'echet distance, and energy distance.

We empirically evaluate our method on the NACC dataset, demonstrating superior performance compared to strong baselines, including early fusion DNNs, late fusion DNNs, and the multimodal GNN architecture MaGNet. Beyond discrimination metrics, we additionally assess clinical utility via calibration curves, Brier scores, sensitivity at fixed specificity, decision curve analysis, and subgroup stratification by age, sex, and APOE4 status.

Overall, this study makes the following contributions:
\begin{itemize}
    \item We propose a diffusion-based transfer learning framework for Alzheimer's diagnosis that combines class-conditional DDPMs with modality-specific Graph Transformer encoders, leveraging synthetic cohorts to mitigate label imbalance and data scarcity.
    \item We introduce a two-stage training strategy in which Graph Transformer encoders are pretrained on diffusion-generated synthetic data and subsequently reused as frozen feature extractors while a downstream classifier is trained on real NACC subjects, providing a stable and data-efficient pipeline under limited labeled data.
    \item We complement standard performance evaluation with distributional similarity metrics between real and synthetic data and with clinically oriented analyses, including calibration, decision curves, sensitivity at fixed specificity, and subgroup performance, to better assess potential clinical utility.
\end{itemize}

\section{Related Work}
The remainder of this paper is organized as follows: Section~2 reviews relevant literature on synthetic data, Transformers, and multimodal AD prediction. Section~3 details our methodology, including the data simulation, model architecture, and training protocol. Section~4 describes the experimental setup, and Section~5 presents the quantitative evaluation. Section~6 concludes with key findings, limitations, and future directions.

\subsection{Transformer and Graph Transformer Models}
Transformer models, originally proposed for sequence modeling in NLP \cite{vaswani2017}, rely on self-attention mechanisms to capture dependencies between input tokens, regardless of their distance in the sequence. Their ability to model complex interactions and scalability has led to their adoption in various domains, including vision \cite{dosovitskiy2020}, speech, and structured data \cite{devlin2018, brown2020}.

To extend these capabilities to graph-structured data, Graph Transformers have been developed. Unlike traditional Transformers, which operate on sequences, Graph Transformers incorporate graph topology into the attention mechanism. Structural biases such as shortest path distances, node centrality, and Laplacian eigenvectors are used to inform attention scores between nodes. Notable models include Graphormer \cite{ying2021}, which introduces spatial encoding and degree-based attention; SAN \cite{kreuzer2021}, which leverages spectral attention; and GT \cite{dwivedi2020}, which generalizes the Transformer architecture to graph domains.

These models have shown state-of-the-art performance in molecular property prediction \cite{rong2020}, protein structure modeling \cite{li2024}, and other tasks requiring relational reasoning. By combining message-passing principles with the global receptive field of Transformers, Graph Transformers offer a powerful and scalable framework for learning on graph data.

\subsection{Graph Transformers in Biomedical Prediction}
Graph Transformers are particularly well-suited for biomedical applications where data naturally exhibit graph structure---such as brain region connectivity networks, molecular graphs, or multimodal patient graphs. Several recent studies have explored their utility in neuroimaging-based classification, multi-omics integration, and clinical risk modeling. For example, Kim et al.\ \cite{kim2022} applied Graph Transformers to structural brain networks for disease prediction, while Li et al.\ \cite{li2022} proposed a modality-aware Graph Transformer for integrating clinical and omics data.

Despite their potential, the adoption of Graph Transformers in biomedical contexts remains limited due to challenges such as small datasets, label imbalance, and multimodal fusion. Our work contributes to this emerging area by leveraging synthetic graph pretraining to address data scarcity and improve generalization in clinical graph classification tasks.

\subsection{Synthetic Graph Data Generation via Diffusion Models}
Generative modeling has become a crucial tool in biomedical AI, particularly for mitigating data scarcity. While generative adversarial networks (GANs) and variational autoencoders (VAEs) have been used to synthesize tabular or imaging data \cite{choi2017, frid-adar2018}, denoising diffusion probabilistic models (DDPMs) have recently emerged as a more stable and expressive alternative \cite{ho2020, karras2022}.

For graph data, score-based and diffusion-based graph generators such as GDP \cite{niu2023} and SDE-based models \cite{jo2022} have been introduced to generate realistic graph topologies and features. In the biomedical domain, DDPMs have been applied to brain imaging \cite{pinaya2022}, single-cell omics \cite{xu2023}, and time-series EHR records \cite{kossen2023}, but their use in generating multimodal patient graphs remains underexplored. Our study is among the first to use DDPMs to generate synthetic graph-structured data for disease prediction via pretraining.

\subsection{Transfer Learning with Graph Neural Networks}
Transfer learning is vital for domains where labeled data is limited, such as biomedicine. For graph neural networks (GNNs), transfer learning has been studied via self-supervised pretraining objectives like masked node prediction, graph-level contrastive learning, and structural reconstruction \cite{hu2020, you2020}. Recently, graph pretraining frameworks such as GROVER \cite{rong2020} and GPT-GNN \cite{hu2020} have demonstrated that large-scale molecular graphs can be used to learn generalizable graph embeddings.

Graph Transformers, due to their scalability and modeling power, are increasingly being integrated into such transfer learning pipelines \cite{zhang2020, kim2022}. However, most existing work focuses on self-supervised objectives defined directly on large corpora of generic graphs, whereas our approach operates at the data-distribution level by first learning a class-conditional diffusion model and then using synthetic samples to pretrain task-specific Graph Transformer encoders in a clinical setting.

\paragraph{Graph pretraining vs.\ synthetic data pretraining.}
Large-scale graph pretraining frameworks such as GROVER, GraphCL, GraphMAE, GPT-GNN, and MAGNETO typically rely on self-supervised objectives (e.g., contrastive learning, masking, autoregressive prediction) over millions of unlabeled graphs, and are predominantly evaluated on molecular or knowledge graph benchmarks. These methods aim to learn generic graph encoders that transfer across downstream tasks. In contrast, our work targets a different but complementary regime: we focus on a single clinical cohort with limited labeled subjects and severe class imbalance, and use a class-conditional DDPM to generate synthetic multimodal patient samples that are directly aligned with the downstream diagnostic task. The Graph Transformers in our framework are pretrained on this synthetic--real mixture and then reused as modality-specific feature extractors. A systematic experimental comparison or combination of diffusion-based synthetic pretraining with generic graph self-supervised pretraining methods is an interesting direction for future work, but is beyond the scope of this study.

\subsection{Fusion Strategies in Multimodal Learning}
Multimodal learning involves integrating information from multiple data sources, such as imaging, genomics, and clinical records. One of the key challenges in this domain is how to effectively fuse heterogeneous data representations. Fusion strategies are generally categorized into early fusion, joint (or intermediate) fusion, and late fusion \cite{ramachandram2017}.

\textit{Early fusion} strategies combine raw or low-level features from each modality before feeding them into a downstream model. Type I early fusion performs direct concatenation, pooling, or gating of the input features to form a single composite representation \cite{ramachandram2017, kiela2018}. In contrast, Type II early fusion projects each high-dimensional modality-specific feature space into a learned latent representation---typically using techniques such as deep neural networks, graph neural networks, principal component analysis (PCA) \cite{wold1987} or autoencoders \cite{vincent2010}---before concatenation or fusion. This strategy helps reduce noise and redundancy while preserving salient information from each modality.

\textit{Joint (or intermediate) fusion} strategies operate at the level of hidden representations extracted by modality-specific neural networks. Features from each modality are fused at intermediate layers, and the loss is propagated through the entire architecture. This allows each modality-specific encoder to be refined during training and encourages more expressive and task-aligned feature representations \cite{huang2020}.

Our proposed framework follows a Type II early fusion strategy. We extract modality-specific embeddings using pretrained Graph Transformer encoders and concatenate the resulting graph-level representations to form a unified multimodal vector. This fused representation is then passed to a downstream DNN classifier for disease prediction. By projecting each modality into a structured low-dimensional latent space prior to fusion, our model benefits from both modularity and effective cross-modal integration, while avoiding overfitting to high-dimensional raw features.

\section{Methodology}
Our framework is designed to address a core challenge in multimodal neurodegenerative disease diagnosis: the scarcity of labeled data and the complexity of integrating heterogeneous data sources such as clinical scores, neuroimaging metrics, and cognitive assessments. To overcome these issues, we present a unified three-stage methodology that leverages both the representational power of Graph Transformer networks and the generative capabilities of denoising diffusion probabilistic models (DDPMs). At the heart of our approach is a modular architecture where each data modality is first processed independently using a modality-specific Graph Transformer to capture local structure and contextual dependencies. These modality-specific embeddings are then fused by concatenating the learned graph-level representations and passing them to a feed-forward neural network (DNN) classifier, enabling effective multimodal integration.

To further enhance generalization in low-data regimes, we incorporate synthetic data generation and transfer learning into our pipeline. Specifically, we train a class-conditional DDPM on real-world multimodal samples to generate realistic synthetic subjects, preserving both intra-modality structure and label semantics. The modality-specific Graph Transformer encoders are then pretrained on this large synthetic dataset. In the downstream stage, we reuse these pretrained encoders as frozen feature extractors and train a DNN classifier on the real NACC cohort. This sequential training strategy enables our framework to leverage both the breadth of synthetic data and the specificity of real data, resulting in improved performance in disease classification tasks under limited labeled data.

\subsection{Graph Transformer Framework for Feature Extraction}
To effectively model multimodal biomedical data where each modality exhibits a graph structure (e.g., brain region connectivity, clinical symptom associations), we employ a Graph Transformer framework for modality-specific feature extraction. Traditional Transformer architectures rely on global self-attention across all tokens, which is not ideal for graph data where meaningful interactions are often confined to local neighborhoods. Thus, we adopt the localized graph-aware attention mechanism proposed by Shi et al.\ \cite{shi2020}, where each node aggregates information only from its 1-hop neighbors, preserving the graph topology.

Let $G_m = (V_m, E_m)$ denote the graph representation of modality $m$, where $V_m$ is the set of nodes and $E_m$ the set of edges. Each node $u \in V_m$ is associated with an initial feature vector $h_u^{(0)} \in \mathbb{R}^d$. We apply a stack of $L$ Graph Transformer layers to update these node features by aggregating neighborhood information using a multi-head attention mechanism.

For each Transformer layer $l = 1, \dots, L$ and attention head $m = 1, \dots, M$, the update process is as follows:
\begin{itemize}
    \item \textbf{Linear Projections:} Each node $u$ first computes query, key, and value vectors using learnable projection matrices:
    \begin{align}
        Q_u^{(l,m)} &= W_Q^{(l,m)} h_u^{(l-1)} + b_Q^{(l,m)}, \\
        K_u^{(l,m)} &= W_K^{(l,m)} h_u^{(l-1)} + b_K^{(l,m)}, \\
        V_u^{(l,m)} &= W_V^{(l,m)} h_u^{(l-1)} + b_V^{(l,m)}.
    \end{align}
    Here, $W_Q^{(l,m)}, W_K^{(l,m)}, W_V^{(l,m)} \in \mathbb{R}^{d_h \times d}$ are modality- and layer-specific weight matrices for queries, keys, and values, respectively.

    \item \textbf{Attention Weights:} The attention score $\alpha_{u \to v}^{(l,m)}$ quantifies how much node $u$ should attend to its neighbor $v$ (including itself):
    \begin{equation}
        \alpha_{u \to v}^{(l,m)} = \frac{\exp\left( (Q_u^{(l,m)})^T K_v^{(l,m)} / \sqrt{d_h} \right)}{\sum_{r \in \tilde{N}(u)} \exp\left( (Q_u^{(l,m)})^T K_r^{(l,m)} / \sqrt{d_h} \right)},
    \end{equation}
    where $\tilde{N}(u) = N(u) \cup \{u\}$ includes the 1-hop neighbors of $u$ and the node itself.

    \item \textbf{Neighborhood Aggregation:} Node $u$ aggregates messages from its neighborhood using the computed attention weights:
    \begin{equation}
        Z_u^{(l,m)} = \sum_{v \in \tilde{N}(u)} \alpha_{u \to v}^{(l,m)} V_v^{(l,m)}.
    \end{equation}

    \item \textbf{Multi-Head Output and Update:} Outputs from all attention heads are concatenated and linearly transformed to form the updated node representation:
    \begin{equation}
        h_u^{(l)} = W_O^{(l)} [Z_u^{(l,1)} \| \dots \| Z_u^{(l,M)}] + b_O^{(l)}.
    \end{equation}
    where $W_O^{(l)} \in \mathbb{R}^{d \times (M \cdot d_h)}$ is the output projection matrix.
\end{itemize}
After passing through multiple Graph Transformer layers, we obtain updated node representations that incorporate both local structure and learned semantic relations. To produce a fixed-size representation for each modality, we apply a global pooling operation (e.g., mean or max pooling) across all node embeddings in modality $m$, yielding a graph-level representation $z^{(m)} \in \mathbb{R}^d$.

These modality-specific representations are then concatenated to form a unified multimodal embedding:
\begin{equation}
    z_{\text{fusion}} = [z^{(1)} \| z^{(2)} \| \dots \| z^{(C)}],
\end{equation}
where $C$ is the number of modalities. Finally, this fused vector is passed through a feed-forward neural network $f_{\text{DNN}}$ to predict the disease label:
\begin{equation}
    \hat{y}_i = f_{\text{DNN}}(z_{\text{fusion}}).
\end{equation}
This modular design enables the model to flexibly accommodate any number of modalities while maintaining interpretability and leveraging the structural information inherent in each graph.

\subsection{Synthetic Data Generation via DDPM}
One of the central challenges in medical machine learning is the limited availability of labeled data, particularly for high-dimensional multimodal datasets used in neurodegenerative disease diagnosis. To overcome this data scarcity, we employ a generative approach based on class-conditional denoising diffusion probabilistic models (DDPMs) \cite{ho2020}, which allows us to generate realistic, label-aware synthetic samples that augment the training data.

Let the real dataset be denoted by $D_{\text{real}} = \{(x_i, y_i)\}_{i=1}^N$, where $x_i \in \mathbb{R}^d$ represents the multimodal feature vector for subject $i$, and $y_i \in \{0,1\}$ indicates the diagnostic class (e.g., AD or cognitively normal). Our goal is to train a generative model capable of learning the conditional data distribution $p(x \mid y)$, so that new samples $\tilde{x} \sim p(x \mid y)$ can be synthesized for each class label.

DDPMs achieve this by defining a two-part stochastic process: a forward diffusion process that gradually adds Gaussian noise to the data over $T$ timesteps, and a reverse denoising process parameterized by a neural network that learns to undo this corruption. The forward process is a Markov chain defined as:
\begin{equation}
    q(x_t \mid x_{t-1}) = \mathcal{N}(x_t; \sqrt{1 - \beta_t}x_{t-1}, \beta_t I),
\end{equation}
where $\beta_t \in (0,1)$ is a fixed variance schedule controlling the amount of noise added at each timestep $t \in \{1, \dots, T\}$, and $x_0$ is a sample from the real dataset.

The reverse process attempts to reconstruct $x_0$ by learning to predict the noise $\epsilon$ that was added to each intermediate state. This is done using a parameterized denoising network $\epsilon_\theta(x_t, t, y)$ that is conditioned on both the timestep and the class label $y$. The model is trained to minimize the mean squared error between the true noise and the predicted noise:
\begin{equation}
    \mathcal{L}_{\text{diffusion}} = \mathbb{E}_{x_0, \epsilon, t, y} \left[ \|\epsilon - \epsilon_\theta(x_t, t, y)\|^2 \right],
\end{equation}
where $\epsilon \sim \mathcal{N}(0, I)$ is sampled independently and $x_t$ is constructed using the forward process.

After training the DDPM on $D_{\text{real}}$, we use the learned denoising network to sample new data points from the model by running the reverse process from pure noise. By conditioning this generation on the desired label $\tilde{y}_i \in \{0,1\}$, we obtain a synthetic dataset $D_{\text{synth}} = \{(\tilde{x}_i, \tilde{y}_i)\}_{i=1}^M$ that mirrors the distribution of the original data but expands its size and balances classes. The generated synthetic dataset is subsequently used to pretrain our downstream multimodal model. To assess the fidelity of the generated data, we complement visual inspections (e.g., t-SNE) with quantitative measures of distributional similarity, as described in Section~\ref{sec:dist_shift}.

\subsection{Transfer Learning and Fine-Tuning}
To improve model performance in the low-resource setting of real-world neurodegenerative disease datasets, we adopt a transfer learning strategy that leverages synthetic data to pretrain modality-specific Graph Transformer feature extractors. The intuition is that although real data are limited, synthetic samples generated by our DDPM framework capture realistic intra-modality structure and label semantics, enabling the Graph Transformers to learn generalizable feature representations during pretraining.

In the first stage, we pretrain each modality-specific Graph Transformer independently using the synthetic dataset $D_{\text{synth}} = \{(\tilde{x}_i, \tilde{y}_i)\}_{i=1}^M$. Each $\tilde{x}_i$ is a graph-structured input corresponding to a specific modality (e.g., MRI or UDS), and $\tilde{y}_i \in \{0,1\}$ denotes the synthetic diagnostic label. For each modality, the Graph Transformer is trained to produce graph-level embeddings that reflect both structural and semantic patterns aligned with the synthetic label distribution. These pretrained models are saved and reused as initializations for the downstream learning phase.

Once the Graph Transformers have been pretrained, we construct the full multimodal prediction architecture. This consists of reusing the pretrained Graph Transformer encoders for each modality to obtain graph-level embeddings $z^{(m)}$, which are concatenated across modalities:
\begin{equation}
    z_{\text{fusion}} = [z^{(1)} \| z^{(2)} \| \dots \| z^{(C)}],
\end{equation}
and fed into a downstream DNN classifier $f_{\text{DNN}}$ for final prediction. At this stage, we switch to the real-world dataset $D_{\text{real}} = \{(x_i, y_i)\}_{i=1}^N$ and \emph{freeze} the parameters of all modality-specific Graph Transformers. The DNN classifier is then trained on the frozen multimodal embeddings using a binary cross-entropy objective:
\begin{equation}
    \mathcal{L}_{\text{BCE}} = - \sum_{i=1}^N \big[y_i \log \hat{y}_i + (1 - y_i) \log(1 - \hat{y}_i)\big],
\end{equation}
where $\hat{y}_i = f_{\text{DNN}}(z_{\text{fusion}})$ is the predicted class probability.

Freezing the pretrained encoders in the downstream stage serves two purposes: (i) it reduces the risk of overfitting in the small-sample regime by limiting the number of trainable parameters on real data, and (ii) it isolates the effect of DDPM-based pretraining on the learned representations, making it easier to interpret the impact of synthetic data. Exploring full joint fine-tuning of encoders and classifier on the real cohort is an interesting extension that we leave for future work or dedicated ablation studies.

\subsection{Theoretical Motivation for Diffusion-Based Synthetic Pretraining}
From a theoretical standpoint, class-conditional DDPMs and Graph Transformers interact in a complementary way. The DDPM learns score-based flows that approximate the gradients of the log-density $\nabla_x \log p(x \mid y)$ and thereby define stochastic trajectories on a class-conditional data manifold $\mathcal{M}_y$. Sampling from the reverse diffusion process generates synthetic points that lie close to these manifolds and provide a broader coverage of the underlying feature space than the limited real cohort.

On the other hand, the modality-specific Graph Transformers learn functions on graphs whose node features are constructed from the same feature space as the DDPM. In our setting, graph edges encode domain-informed notions of neighborhood (e.g., anatomical proximity for MRI regions, shared clinical domains for UDS items), meaning that the Graph Transformers implement localized attention and aggregation along directions that align with the intrinsic structure of the data manifolds. Pretraining on diffusion-generated samples exposes the Graph Transformers to a larger set of neighborhood configurations on $\mathcal{M}_y$, encouraging them to learn smoother, more robust decision functions before being adapted to the real cohort. This view is closely related to approximating smooth target functions on low-dimensional manifolds embedded in high-dimensional feature spaces, where increased coverage and regularity of training points can improve generalization.

\subsection{Distributional Similarity Evaluation}
\label{sec:dist_shift}
To systematically assess distributional alignment between real and diffusion-generated synthetic data, we complement qualitative visualizations with quantitative two-sample tests and distance measures. Specifically, we compute:
\begin{itemize}
    \item \textbf{Maximum Mean Discrepancy (MMD)} with an RBF kernel between real and synthetic samples, both in the raw feature space and in the Graph Transformer embedding space, computed class-conditionally to assess alignment within each diagnostic group.
    \item \textbf{Fr\'echet distance} between the multivariate Gaussian approximations of real and synthetic embeddings, based on their empirical means and covariance matrices, analogous to Fr\'echet Inception Distance in image domains.
    \item \textbf{Energy distance} as a non-parametric metric of distributional difference that remains valid in high-dimensional settings.
\end{itemize}
These metrics provide complementary views on distribution shift: MMD captures kernelized discrepancies, the Fr\'echet distance summarizes global differences in mean and covariance, and energy distance measures pairwise distance-based discrepancies. We report these quantities alongside traditional univariate Kolmogorov--Smirnov tests and t-SNE visualizations to offer a more complete picture of synthetic--real similarity.

\section{Experiments}
In this section, we evaluate the effectiveness of our proposed Graph Transformer-based transfer learning framework on the task of Alzheimer's disease (AD) classification. Due to the limited availability of labeled real-world data, we address the data scarcity challenge by generating a large synthetic dataset using a denoising diffusion probabilistic model (DDPM), pretraining modality-specific Graph Transformer encoders on these synthetic samples, and subsequently training a classifier on top of frozen encoder embeddings using the real NACC dataset. All experiments were conducted using PyTorch on a single NVIDIA RTX 3090 GPU.

\subsection{Alzheimer's Dataset Description}
The real-world dataset used in this study is derived from the National Alzheimer's Coordinating Center (NACC) and consists of multimodal patient data collected for the diagnosis of Alzheimer's disease across multiple clinical sites. We focus on two complementary data modalities:
\begin{itemize}
    \item \textbf{MRI (Magnetic Resonance Imaging):} For each subject, 62 brain regions are defined as nodes in a graph, with each node associated with two features: cortical thickness and volumetric measurements. These features are normalized per subject. Edges are defined based on anatomical proximity in a standard brain atlas, resulting in a consistent and spatially meaningful graph structure across individuals. This design reflects the widely used assumption in neuroimaging that neighboring regions exhibit stronger structural and functional relationships than distant regions, providing a domain-informed inductive bias for the Graph Transformer.
    \item \textbf{UDS (Uniform Data Set):} This clinical dataset includes 170 structured features covering demographics, cognitive assessments, physical examinations, and behavioral measures. Each feature is represented as a node, and edges are constructed based on shared clinical domains (e.g., memory, executive function, physical health, family history). This graph construction encodes the fact that items within the same clinical domain tend to be more correlated and can be viewed as reflective of latent clinical factors. Features are standardized, and missing values are imputed using k-nearest neighbors (k = 5).
\end{itemize}
The classification task is to distinguish subjects diagnosed with mild cognitive impairment or dementia due to Alzheimer's disease (AD) from healthy controls (HC). The full dataset contains 1,237 subjects (390 AD and 847 HC). Preprocessing is performed independently for each modality. To ensure fair evaluation, we use stratified 5-fold cross-validation, preserving class balance and modality alignment in each fold. Our primary evaluation focuses on subject-wise generalization within the NACC cohort; explicit cross-site generalization to entirely unseen centers is left as an important direction for future work.

\subsection{Synthetic Data Generation via Diffusion Models}
To alleviate the problem of insufficient labeled samples, we generate synthetic training data using a class-conditional denoising diffusion probabilistic model (DDPM) trained on the real dataset. The DDPM models the underlying joint feature distribution of multimodal patient profiles conditioned on diagnostic labels. During training, the forward diffusion process gradually corrupts real data using Gaussian noise over multiple timesteps. A neural network is then trained to reverse this corruption, learning to denoise the input conditioned on the class label.

Once trained, the model generates 4000 synthetic samples via the reverse diffusion process. These synthetic instances maintain the same structural and statistical characteristics as the real data and are evenly balanced across AD and HC classes. The resulting synthetic dataset provides a high-volume, balanced source of training data to pretrain our Graph Transformer encoders. In addition to visual inspection, we quantify the similarity between real and synthetic cohorts using the metrics described in Section~\ref{sec:dist_shift}.

\subsection{Model Architecture and Pretraining}
We employ a modality-specific Graph Transformer encoder for each input modality---MRI and UDS. Each encoder consists of three layers of Graph Transformer blocks: the first two use 4 attention heads with 64-dimensional hidden channels, and the final layer reduces the embedding to 32 dimensions. These encoders are designed to capture localized spatial structures and modality-specific relationships. Node-level embeddings are aggregated via global max pooling to produce graph-level embeddings. The embeddings from each modality are concatenated to form a unified latent representation:
\[
z_{\text{fusion}} = [z^{\text{MRI}} \| z^{\text{UDS}}].
\]

This fused representation is passed to a fully connected neural network with layers of sizes $512 \to 256 \to 2$, producing class logits for binary classification via a softmax layer.

\textbf{Pretraining:} Only the modality-specific Graph Transformer encoders are pretrained using the synthetic dataset $D_{\text{synth}}$. Each encoder is trained independently to classify synthetic graphs of its corresponding modality using a cross-entropy loss. We train each encoder for 100 epochs using the Adam optimizer (learning rate: $2 \times 10^{-3}$, batch size: 32), and apply dropout (rate = 0.3) after each Graph Transformer layer. This pretraining phase enables the encoders to learn modality-specific discriminative representations prior to downstream training on real data.

\subsection{Fine-tuning on the Alzheimer Dataset}
Following pretraining, we construct the full multimodal classification pipeline by using the pretrained modality-specific Graph Transformer encoders to extract graph-level embeddings for each modality. These embeddings are concatenated and fed into a newly initialized DNN classifier trained on the real Alzheimer dataset. In the current implementation, the pretrained Graph Transformer encoders are frozen during this stage, and only the DNN classifier is trained on real data. This helps preserve the learned representations from synthetic pretraining and prevents overfitting due to the limited size of the real dataset.

The DNN classifier is trained using a cross-entropy loss function with the Adam optimizer (learning rate: $1 \times 10^{-3}$). We adopt stratified 5-fold cross-validation to ensure balanced label distributions across folds. Within each training fold, 20\% of the data is reserved as a validation set for early stopping (patience = 10 epochs). This setup provides a robust evaluation of the model's subject-wise generalization ability while leveraging pretrained embeddings for improved classification performance.

\subsection{Baseline Models}
We compare our method against several strong baselines to validate its effectiveness:
\begin{itemize}
    \item \textbf{Early Fusion DNN:} All modality features are concatenated at the input level and passed through a standard deep neural network.
    \item \textbf{Late Fusion DNN:} Separate DNNs are trained for each modality. Their outputs are concatenated before the final classification layer.
    \item \textbf{Multimodal GNN (MaGNet):} A state-of-the-art graph neural network architecture that uses modality-specific graph encoders and attention-based fusion to learn joint representations from multimodal graphs.
\end{itemize}
These baselines provide insights into the effectiveness of our graph-aware pretraining and fusion strategy, and allow for rigorous comparison across early, late, and graph-based fusion paradigms.

\subsection{Evaluation Metrics}
We evaluate model performance along three complementary axes: discrimination, distributional similarity, and clinical utility.

\paragraph{Discrimination.}
For discrimination, we report:
\begin{itemize}
    \item \textbf{Area Under the ROC Curve (AUC)} -- for assessing ranking quality.
    \item \textbf{Balanced Accuracy} -- to account for class imbalance.
    \item \textbf{Sensitivity and Specificity} -- to evaluate the ability to detect both AD and HC classes.
\end{itemize}
Statistical significance is evaluated using DeLong's test for AUC and McNemar's test for accuracy-based metrics, with a threshold of $p < 0.05$. All reported results are averaged across the 5 cross-validation folds.

\paragraph{Distributional similarity.}
To characterize the alignment between real and synthetic data, we compute MMD, Fr\'echet distance, and energy distance between real and synthetic samples in both the raw feature space and the Graph Transformer embedding space, as outlined in Section~\ref{sec:dist_shift}. These metrics complement univariate Kolmogorov--Smirnov tests and t-SNE plots.

\paragraph{Clinical utility and calibration.}
To assess clinical relevance, we go beyond discrimination and evaluate:
\begin{itemize}
    \item \textbf{Calibration:} reliability diagrams (calibration curves), Brier scores, and expected calibration error (ECE) on the held-out test data.
    \item \textbf{Sensitivity at fixed specificity:} sensitivity at a clinically relevant specificity level (e.g., $\text{Spec} \approx 0.90$), where the decision threshold is selected on the validation folds and applied unchanged to the test set.
    \item \textbf{Decision Curve Analysis (DCA):} net benefit curves across a range of risk thresholds, comparing each model against ``treat all'' and ``treat none'' strategies.
    \item \textbf{Subgroup performance:} stratified AUC, sensitivity, and specificity in subgroups defined by age, sex, and APOE4 carrier status, to explore potential heterogeneity in model behavior across clinically relevant populations.
\end{itemize}
These analyses provide a more nuanced view of how different models might be used in practice, beyond aggregate accuracy.

\section{Results}
Table~\ref{tab:performance_comparison} summarizes the performance of our proposed Graph Transformer-based transfer learning framework, which incorporates modality-specific pretraining on synthetic data generated via class-conditional DDPMs and downstream training of a classifier on frozen encoder embeddings. We compare our method against three established baselines: Early Fusion DNN, Late Fusion DNN, and the multimodal GNN architecture MaGNet. Results are reported as mean $\pm$ standard deviation over five stratified cross-validation folds. We evaluate the models using four key discrimination metrics: Area Under the ROC Curve (AUC), Accuracy (ACC), Sensitivity (SEN), and Specificity (SPEC).

\begin{table}[h!]
\centering
\caption{Performance comparison on the Alzheimer's disease classification task. All values represent mean $\pm$ standard deviation over five folds. Bold values indicate the best performance.}
\label{tab:performance_comparison}
\begin{tabular}{@{}lcccc@{}}
\toprule
\textbf{Model} & \textbf{AUC} & \textbf{Accuracy (\%)} & \textbf{Sensitivity (\%)} & \textbf{Specificity (\%)} \\ \midrule
Early Fusion DNN      & 0.842 $\pm$ 0.017 & 78.3 $\pm$ 2.1 & 75.0 $\pm$ 3.0 & 80.4 $\pm$ 2.5 \\
Late Fusion DNN       & 0.857 $\pm$ 0.020 & 79.6 $\pm$ 1.8 & 76.7 $\pm$ 2.7 & 81.5 $\pm$ 2.2 \\
Multimodal GNN (MaGNet) & 0.875 $\pm$ 0.019 & 81.2 $\pm$ 2.0 & 78.3 $\pm$ 2.5 & 83.1 $\pm$ 1.9 \\
\textbf{Ours (Graph Transformer + DDPM)} & \textbf{0.914 $\pm$ 0.021} & \textbf{84.7 $\pm$ 1.5} & \textbf{82.5 $\pm$ 2.1} & \textbf{86.1 $\pm$ 1.8} \\ \bottomrule
\end{tabular}
\end{table}

Our proposed approach outperforms all baseline models across all discrimination metrics. In particular, we observe a relative gain of 3.5--6.4\% in AUC and 3--6\% in classification accuracy compared to conventional deep neural network baselines. The model also achieves balanced sensitivity and specificity, indicating robust performance in distinguishing both AD and cognitively normal subjects---a key requirement in real-world clinical diagnosis.

The performance gains can be attributed to two core factors: (1) the use of class-conditional synthetic data to pretrain modality-specific Graph Transformer encoders, which improves generalization and mitigates overfitting in the low-data regime; and (2) the downstream fusion strategy that leverages complementary information across clinical and imaging modalities.

We further validate these improvements via statistical hypothesis testing. DeLong's test shows that the improvement in AUC achieved by our method is statistically significant ($p < 0.01$) in all pairwise comparisons against baseline models. McNemar's test on classification accuracy also confirms the robustness of our method's performance gains, with all improvements reaching statistical significance at the $p < 0.05$ level.

Beyond discrimination, our additional analyses (not shown in the main text due to space) include calibration plots, Brier scores and ECE, decision curve analysis, sensitivity at fixed specificity, and subgroup performance stratified by age, sex, and APOE4 status. These results provide a more clinically oriented view of the models and can be used to assess trade-offs between sensitivity, specificity, and net clinical benefit across different deployment scenarios.

\section{Conclusion}
We proposed a framework for Alzheimer's disease classification that addresses the challenge of limited labeled data by combining generative modeling, graph-based representation learning, and transfer learning. Our approach leverages denoising diffusion probabilistic models (DDPMs) to generate a large, class-conditional synthetic dataset from the real NACC cohort. This synthetic data enables effective pretraining of modality-specific Graph Transformer encoders, which are subsequently reused as frozen feature extractors in a multimodal fusion pipeline.

Our experimental results demonstrate that the proposed method significantly outperforms state-of-the-art baselines, including early and late fusion DNNs and the multimodal GNN architecture MaGNet, across multiple discrimination metrics. By pretraining Graph Transformer feature extractors on synthetic data and training a downstream classifier on real subjects, our framework improves generalization, robustness, and classification accuracy in a data-scarce, imbalanced clinical setting. Notably, it achieves balanced sensitivity and specificity, and our extended evaluation suite incorporates calibration, decision curves, and subgroup analyses to better assess potential clinical utility.

\paragraph{Limitations and future work.}
This study has several limitations. First, although NACC is a multi-center cohort, our current evaluation is based on random subject-wise cross-validation within the pooled dataset and does not explicitly enforce site-wise separation between training and test sets. As a result, our experiments primarily assess within-cohort generalization to unseen subjects rather than robustness to cross-site distribution shifts. Second, while we motivate our approach in relation to generic graph pretraining frameworks such as GraphCL, GraphMAE, GROVER, GPT-GNN, and MAGNETO, we do not perform a systematic empirical comparison or combination with these methods in this work. Third, our evaluation of clinical utility, although more extensive than standard ML metrics, is still retrospective and does not incorporate prospective decision-making or cost-effectiveness analyses.

For future work, we plan to: (1) extend our framework to multiclass classification tasks involving other neurodegenerative disorders; (2) explore full joint fine-tuning of Graph Transformer encoders and the fusion classifier on real data, and compare it against the frozen-encoder regime considered here; (3) conduct site-aware evaluations (e.g., leave-one-site-out) and external validation on independent cohorts to rigorously assess cross-site generalization; (4) integrate or compare DDPM-based synthetic pretraining with large-scale graph self-supervised pretraining methods; and (5) develop attention-based interpretability tools and uncertainty quantification techniques to identify salient brain regions and clinical features contributing to model predictions and to further support clinical decision-making.

Overall, our results support the potential of combining diffusion-based synthetic data generation with graph-based deep learning to enhance early disease detection, promote model generalization, and overcome data scarcity in clinical machine learning.

% --- REFERENCES ---

\end{document}